\documentclass{trbunofficial}
\usepackage{amsmath}
\usepackage{array,booktabs,longtable}
\newcolumntype{L}[1]{>{\raggedright\arraybackslash}p{#1}}

\begin{document}


\title{Mitigating Bus Bunching with Reinforcement Learning Enhanced by Semantic Stop Embedding}
\TRBauthor{Xin Dong\textsuperscript{*}}{Department of Civil and Environmental Engineering, The Pennsylvania State University}{xjd5036@psu.edu}[University Park, PA, 16803][0009-0000-9420-6496]
\TRBauthor{Vikash V. Gayah}{Department of Civil and Environmental Engineering, The Pennsylvania State University}{gayah@engr.psu.edu}[University Park, PA, 16803][0000-0002-0648-3360]

\TRBtitlefootnote{\textsuperscript{*} Corresponding author}

\AuthorHeaders{Dong, Gayah}


\maketitle

\section{Abstract}
Bus bunching degrades service regularity and increases passenger waiting in high-frequency transit. Existing reinforcement-learning-based holding controllers primarily rely on instantaneous operational variables or route-specific stop identifiers, which provide limited information about the functional and operational context of individual stops and constrain policy reuse across routes. This study introduces an LLM-assisted semantic stop representation for event-driven bus holding control. An LLM is used offline to transform heterogeneous stop information, including physical attributes, surrounding activity context, and historical operational characteristics, into fixed semantic embeddings that are incorporated into a deep Q-learning controller without requiring real-time LLM inference. Experiments are conducted in stochastic simulations calibrated with observed data from two bus routes. Compared with the best calibrated Daganzo baseline, the semantic controller reduces headway variability, bunching events, and passenger waiting time by 32.0\%, 69.2\%, and 24.0\%, respectively. A route-specific stop identifier does not improve the spacing-only controller, whereas semantic stop information improves headway regularity, waiting time, and holding effort, providing a more favorable overall trade-off across control objectives. Cross-route experiments further show that zero-shot transfer provides limited immediate generalization, while warm-start fine-tuning accelerates early-stage learning and improves transferred policies; cold-start training nevertheless achieves the best final performance. These findings suggest that semantic state representations can complement conventional operational states and support adaptation-based policy reuse across related transit routes. \\
\textbf{Key Words:} Reinforcement learning; semantic stop embeddings; large language models; transfer learning; bus holding control.

\newpage

\section{Introduction}\label{sec:intro}
Bus bunching is a persistent problem in high-frequency transit. Small disturbances in travel time, dwell time, passenger arrivals, traffic conditions, signals, or driver behavior can disrupt vehicle spacing and trigger a self-reinforcing process \cite{DaganzoPilachowski2011Cooperation}. A delayed bus encounters more passengers and experiences longer dwell times, while the following bus serves fewer passengers and gradually closes the gap \cite{DaganzoPilachowski2011Cooperation}. These dynamics can amplify modest headway deviations into persistent bunching, resulting in longer and less predictable passenger waits, uneven vehicle loads, and increased in-vehicle delay \cite{CatsEtAl2011Holding}. At the system level, bunching also reduces service reliability, effective fleet capacity, and operational productivity. Bus-control literature includes schedule- and headway-based holding, stop-skipping, boarding limits, speed control, signal priority, short turning, and combinations of these measures. Holding remains especially attractive because it can be implemented with existing vehicle-location and driver-communication systems.

Classical bus holding control has largely been developed through analytical and rule-based formulations that translate observed headway deviations into real-time holding decisions. \citet{Daganzo2009Headway} derived an adaptive headway-based rule designed to contain local disturbances without requiring substantial schedule slack, while \citet{DaganzoPilachowski2011Cooperation} showed that two-way spacing information and bus-to-bus cooperation can stabilize service. Subsequent studies demonstrated that the effectiveness of such strategies depends on the holding criterion and control-point configuration \cite{CatsEtAl2011Holding}. An empirical comparison by \citet{BerrebiEtAl2018Comparing} found that schedule-based methods required relatively little holding but provided limited headway stabilization, whereas prediction-based methods achieved a better regularity--holding tradeoff but were sensitive to prediction accuracy. Recent reviews further identify passenger-oriented objectives, coordinated control, simplified demand and network assumptions, and limited real-world evaluation as continuing research gaps \cite{GkiotsalitisCats2021AtStop,RezazadaEtAl2024Review}. Although feedback and rule-based strategies provide interpretable and computationally tractable control, their decisions are constrained by predefined functional forms and calibrated parameters, limiting their ability to represent complex operational tradeoffs and adapt to changing system conditions.
A parallel stream of research has formulated bus holding as a real-time optimization problem in which passenger costs, vehicle operations, and service constraints are represented explicitly. \citet{EberleinEtAl2001Holding} developed a rolling-horizon formulation using real-time vehicle information, while \citet{KoehlerEtAl2011IQP} introduced an iterative quadratic procedure for control at multiple holding points. \citet{DelgadoEtAl2012Holding} further incorporated vehicle-capacity constraints, passenger boarding processes, and joint holding and boarding control. Later work compared deterministic and stochastic predictive-control formulations under different operating conditions \cite{MunozEtAl2013Dynamic}, extended holding optimization to corridors with interacting bus services \cite{HernandezEtAl2015MultiLine}, and incorporated dynamically varying passenger demand and link travel times \cite{SanchezMartinezEtAl2016Dynamic}. These models rely on explicit representations of bus and passenger dynamics and require repeated online optimization as new information becomes available. Although existing solution procedures can support real-time implementation, their computational requirements may increase substantially with the prediction horizon and system complexity.

Reinforcement-learning-based control shifts the computationally intensive process of policy optimization to pre-deployment training \cite{liu2021privacy, hou2026mobility}. Once trained, the RL-based controller can generate state-dependent holding actions during operation through a forward pass of the policy network, without repeatedly solving an online optimization problem. Early studies introduced deep value-based learning for discrete holding decisions \cite{AlesianiGkiotsalitis2018RL} and combined approximate dynamic programming with Q-learning to evaluate downstream effects over multiple stages \cite{HeEtAl2022Qlearning}. Later work formulated fleet-wide holding as an event-driven decision process \cite{ZhangEtAl2026SingleAgent} and developed coordinated multi-agent controllers for asynchronous decisions and inter-vehicle interactions \cite{ChenEtAl2016MARL}. Subsequent extensions considered joint holding and stop-skipping \cite{RodriguezEtAl2023Cooperative}, robust and multi-objective control \cite{WangSun2023Robust,WangSun2023MultiObjective}, hierarchical action selection \cite{YuEtAl2024Hierarchical}, and graph-based coordination in congested multi-line corridors \cite{LiEtAl2026GraphAware}.
Despite these advances, existing state representations largely rely on operational variables, such as headways, vehicle loads and passenger occupancy \cite{wood2023development}, passenger flows, previous actions, and categorical vehicle or stop identifiers. These features describe the current fleet state, but categorical identifiers provide little information about the functional and land-use context of individual stops or the relationships between stops across routes. Recent evidence shows that state and action design can matter as much as the selected RL algorithm. In particular, a systematic benchmark found that forward and backward spacing alone can outperform a richer state containing a stop identifier, while discrete holding actions can perform comparably to continuous actions \cite{XuEtAl2025Systematic}. This result does not imply that stops are operationally interchangeable; rather, it suggests that an arbitrary identifier is a weak representation of the contextual factors that distinguish them. This representation gap suggests a potential role for LLMs, whose application in bus holding control remains limited. Existing bus control work has primarily used LLMs to generate and refine reward functions \cite{YuEtAl2025LLMReward}, rather than to construct semantic state representations. Whether LLM-derived stop semantics can improve within-route control and support cross-route policy adaptation therefore remains insufficiently explored.

To summarize, three key gaps remain in existing RL-based bus holding and LLM-enhanced control studies: (1) conventional state representations provide limited information about stop-level functional context; (2) route-specific representations constrain policy reuse across routes; and (3) existing LLM applications mainly focus on reward design rather than semantic state representation. To address these gaps, this study introduces an LLM-assisted semantic state representation for RL-based bus holding control. Rather than treating stops as arbitrary route-specific identifiers, the proposed approach represents their functional and operational characteristics in a shared feature space that can be used across routes. The LLM is employed offline to construct the stop representations, while online holding decisions remain governed by the learned RL policy, avoiding real-time LLM inference. The study investigates whether semantic stop information can improve holding control within a route and facilitate policy reuse across routes. In particular, it examines both the immediate generalization of a pretrained policy and its ability to adapt to a new operating environment through further training. These questions are evaluated in stochastic, data-calibrated bus operation environments through comparisons with classical control strategies and alternative RL state representations.

The main contributions of this study are as follows:

\begin{itemize}
\item We introduce an LLM-assisted semantic representation of bus stops for reinforcement learning-based holding control. The representation combines functional and access-related labels derived from geospatial context with discretized historical operational characteristics, and encodes them in a common continuous feature space across routes. The stop representations are constructed offline and reused throughout RL training and deployment, avoiding repeated and computationally costly semantic feature construction.

\item We use controlled within-route comparisons to isolate the contribution of semantic stop information from that of route-specific stop identity and directly observed passenger-flow variables. The analysis demonstrates the added value of semantic representations for selected control settings and clarifies their role beyond conventional spacing-based state designs.

\item We evaluate cross-route policy reuse through zero-shot transfer, target-route fine-tuning, and cold-start training. The findings reveal that target-route adaptation substantially improves transferred policies and early-stage learning, highlighting the distinction between immediate generalization and adaptation-based reuse.

\end{itemize}

The remainder of this paper is organized as follows. Section~\ref{sec:method} presents the proposed methodology, including the event-driven MDP formulation, semantic stop embedding construction, and semantic-enhanced RL holding policy. Section~\ref{sec:experiment-setup} describes the experimental design. Section~\ref{sec:results} reports and discusses the same-route and cross-route experimental results. Finally, Section~\ref{sec:conclusion} concludes the paper and discusses its limitations and directions for future research.

\section{Methodology}
\label{sec:method}

This section presents a semantic-enhanced reinforcement learning framework for bus holding control. At each bus-departure event, a shared DQN selects a discrete holding duration based on a local observation that combines the bus's spatial relationship with its neighboring buses and a fixed semantic embedding of the current stop. The framework consists of three components: an event-driven Markov decision process formulation, an offline semantic stop embedding procedure, and a DQN-based policy-learning method.

\subsection{Event-Driven MDP Formulation}
\label{subsec:mdp-formulation}

Consider a bus route represented as a directed loop $\mathcal{G}=(\mathcal{S},\mathcal{E})$, where $\mathcal{S}=\{1,\ldots,N\}$ is the set of stops and $\mathcal{E}=\{(i,i+1):i=1,\ldots,N\}$ is the set of consecutive links, with stop $N+1$ identified as stop $1$. Let $\ell_i$ denote the length of link $i$ and $L=\sum_{i=1}^{N}\ell_i$ denote the total loop length. A fleet $\mathcal{B}=\{1,\ldots,B\}$ operates continuously along the route.
The complete system state at time $t$, denoted by $X(t)\in\mathcal{X}$, contains the information required to characterize the bus operation process, including bus positions, onboard passenger loads by destination, stop-level passenger queues, and the service, holding, or travel status of each bus. System transitions are governed by stochastic passenger arrivals, passenger service, link travel times, and holding actions. 
Holding decisions are made asynchronously at bus-departure events. Let $\tau_n^b$ denote the $n$th decision epoch of bus $b$, defined as the time at which the bus completes passenger service at stop $i_n^b$ and is ready to enter the downstream link. At $\tau_n^b$, the controller receives a local observation $\mathbf{o}_n^b=\Omega(X(\tau_n^b))$, selects an action $a_n^b$, and receives the corresponding reward when the same bus reaches its next decision epoch $\tau_{n+1}^b$. The problem is formulated as an event-driven, partially observed Markov decision process:

\begin{equation}
\mathcal{M}=(\mathcal{X},\mathcal{O},\mathcal{A},P,R,\gamma,\Omega).
\label{eq:mdp-formulation}
\end{equation}

Here, $\mathcal{X}$ and $\mathcal{O}$ denote the full-state and observation spaces, respectively; $\mathcal{A}$ is the holding-action space; $P$ denotes the transition dynamics between consecutive decision events; $R$ is the reward function; $\gamma$ is the discount factor; and $\Omega$ maps the full system state to the local controller observation. Although the elapsed time between consecutive decision epochs is variable, each bus-departure event is treated as one decision step by the RL controller.

\subsubsection{State}
\label{subsec:rl-state}

The dynamic component of the observation describes the spatial relationship between the controlled bus and its nearest neighboring buses. Let $x_b(\tau_n^b)\in[0,L)$ denote the longitudinal position of bus $b$ along the loop at decision epoch $\tau_n^b$. The normalized forward and backward spatial spacings are defined as: 

\begin{align}
\bar g_n^{+,b} &= \frac{1}{L}\min_{b'\in\mathcal{B}\setminus\{b\}}\left[\bigl(x_{b'}(\tau_n^b)-x_b(\tau_n^b)\bigr)\bmod L\right], \\
\bar g_n^{-,b} &= \frac{1}{L}\min_{b'\in\mathcal{B}\setminus\{b\}}\left[\bigl(x_b(\tau_n^b)-x_{b'}(\tau_n^b)\bigr)\bmod L\right].
\label{eq:spatial-spacings}
\end{align}

These quantities measure the route distances from the controlled bus to the nearest bus ahead and behind, respectively. Equal forward and backward spacings indicate that the controlled bus is locally centered between its neighboring buses.

Let $\mathbf z_{i_n^b}\in\mathbb R^d$ denote the fixed semantic embedding of the stop at which the decision is made. The proposed semantic-enhanced observation is

\begin{equation}
\mathbf{o}_{n,\mathrm{sem}}^b=\left[\bar g_n^{+,b},\bar g_n^{-,b},\mathbf z_{i_n^b}\right]\in\mathbb R^{2+d}.
\label{eq:semantic-observation}
\end{equation}

The semantic vector provides contextual information that cannot be inferred directly from instantaneous bus spacing, including the physical, functional, and operational characteristics of the current stop. Its construction is described in Section~\ref{subsec:semantic-stop-info}.

Two alternative observation designs are used as benchmarks: the spacing-only observation excludes stop-level contextual information; and the stop-ID observation supplements the spacing variables with the categorical index $i_n^b$ of the current stop:

\begin{equation}
\mathbf{o}_{n,\mathrm{sem}}^b=\left[\bar g_n^{+,b},\bar g_n^{-,b},\mathbf i_n^b\right]\in\mathbb R^{3}.
\label{eq:stop-id-observation}
\end{equation}



\subsubsection{Action}
\label{subsec:rl-action}

At decision points, the controller selects an action index from $\mathcal{A}=\{0,\ldots,A-1\}$. Let $H_{\max}$ denote the maximum allowable holding time. The holding duration associated with action $a\in\mathcal{A}$ is

\begin{equation}
H(a)=\frac{a}{A-1}H_{\max}.
\label{eq:holding-action}
\end{equation}


\subsubsection{Reward}
\label{subsec:rl-reward}

The reward for action $a_n^b$ is assigned when bus $b$ reaches its next decision epoch $\tau_{n+1}^b$. It combines a spacing-equalization term, a penalty for newly detected bunching conditions, and a regularization term favoring shorter holding durations.

Let $C(t)$ denote the cumulative number of bunching conditions detected by time $t$. The three reward components are defined as
\begin{align}
R_{n+1}^{\mathrm{eq},b} &= \exp\!\left(-\left|\bar g_{n+1}^{+,b}-\bar g_{n+1}^{-,b}\right|\right), \label{eq:equalization-reward}\\
\Delta C_{n+1}^b &= \max\!\left\{0,C(\tau_{n+1}^b)-C(\tau_n^b)\right\}, \label{eq:bunching-increment}\\
R_n^{\mathrm{hold},b} &= \exp\!\left(-\frac{H(a_n^b)}{H_{\max}}\right). \label{eq:holding-reward}
\end{align}

The final reward is
\begin{equation}
r_{n+1}^b=R_{n+1}^{\mathrm{eq},b}-\kappa\Delta C_{n+1}^b+\omega R_n^{\mathrm{hold},b},
\label{eq:rl-reward}
\end{equation}
where $\kappa>0$ and $\omega>0$ determine the relative weights assigned to bunching prevention and holding reduction. Their values are reported in Section~\ref{sec:evaluation-protocol}. 


\subsection{Semantic Stop Embedding}
\label{subsec:semantic-stop-info}

To augment the dynamic operating state with stop-level contextual information, we construct a fixed semantic embedding $\mathbf z_i$ for each stop $i$. The embedding is generated offline, cached before RL training, and remains unchanged during policy learning and evaluation. Its construction consists of two stages: constrained semantic annotation of heterogeneous stop information and projection of the resulting representation into a shared low-dimensional feature space. 


\subsubsection{Fact Assembly and Semantic Annotation}
\label{subsec:semantic-annotation}

Let $\boldsymbol{\xi}_i$ denote the structured fact record associated with stop $i$. The record contains three categories of information: physical stop attributes, surrounding activity context, and operational descriptors. Physical attributes characterize the stop facility and its local spatial configuration. Surrounding activity context describes nearby land-use functions and their approximate accessibility, while operational descriptors summarize recurrent passenger-demand and service characteristics.

A deterministic prompt-construction function $\Pi(\cdot)$ converts the structured record into a constrained prompt. A large language model $G_{\psi}$ maps the supplied facts to structured semantic metadata, after which a deterministic normalization function $\mathcal N(\cdot)$ enforces the predefined schema and controlled vocabularies:

\begin{equation}
\mathbf y_i=\mathcal N\!\left(G_{\psi}\!\left(\Pi(\boldsymbol{\xi}_i)\right)\right)=\left(s_i,\mathcal T_i^{\mathrm{role}},r_i,\mathcal T_i^{\mathrm{demand}},\mathcal T_i^{\mathrm{quality}}\right).
\label{eq:semantic-annotation}
\end{equation}

Here, $s_i$ is a short natural-language summary, $\mathcal T_i^{\mathrm{role}}$ is a set of functional role tags, $r_i$ is a route-context category, $\mathcal T_i^{\mathrm{demand}}$ contains demand-related tags, and $\mathcal T_i^{\mathrm{quality}}$ contains data-quality indicators. 

The free-text summary $s_i$ is retained for inspection but is not included in the embedding input. Instead, a leakage-controlled subset of the original attributes, denoted by $\bar{\boldsymbol{\xi}}_i$, is combined with the normalized semantic labels through a deterministic key--value serialization:

\begin{equation}
t_i=S\!\left(\bar{\boldsymbol{\xi}}_i,\mathcal T_i^{\mathrm{role}},r_i,\mathcal T_i^{\mathrm{demand}},\mathcal T_i^{\mathrm{quality}}\right).
\label{eq:semantic-serialization}
\end{equation}

To reduce route-specific leakage, the serialized embedding input excludes explicit route names and raw stop-sequence indices. Continuous or highly variable attributes are represented using predefined categorical bins rather than raw values. Exact geographic coordinates are used only to retrieve nearby contextual information and are not directly embedded. The retained fields are selected to describe stop characteristics that can be compared across routes, including facility type, nearby activity categories, accessibility, demand patterns, service variability, and observation confidence.

\subsubsection{Embedding Construction and Alignment}
\label{subsec:semantic-embedding-alignment}

The canonical text representation $t_i$ is mapped to a high-dimensional vector using a pretrained text-embedding model $E_{\phi}$:

\begin{equation}
\mathbf e_i=E_{\phi}(t_i)\in\mathbb R^{D}.
\label{eq:raw-stop-embedding}
\end{equation}

Because the raw text embedding is substantially larger than the dynamic operational state, it is standardized and projected into a compact feature space before being supplied to the RL controller. Let $\mathcal U$ denote the set of stops used to fit the representation transformation, and let $\boldsymbol{\mu}_e$ and $\boldsymbol{\sigma}_e$ denote the featurewise mean and standard deviation of their raw embeddings. The standardized embedding is

\begin{equation}
\widetilde{\mathbf e}_i=(\mathbf e_i-\boldsymbol{\mu}_e)\oslash\max(\boldsymbol{\sigma}_e,\varepsilon),
\label{eq:embedding-standardization}
\end{equation}

where $\oslash$ denotes elementwise division and $\varepsilon>0$ ensures numerical stability.

Let $\mathbf V_d\in\mathbb R^{D\times d}$ contain the leading $d$ principal-component loading vectors fitted from the standardized embeddings. The reduced representation is

\begin{equation}
\mathbf p_i=\mathbf V_d^{\mathsf T}\widetilde{\mathbf e}_i\in\mathbb R^d.
\label{eq:pca-stop-embedding}
\end{equation}

The PCA coordinates are standardized again and clipped to limit extreme feature values:

\begin{equation}
\mathbf z_i=\operatorname{clip}_{[-c,c]}\!\left((\mathbf p_i-\boldsymbol{\mu}_p)\oslash\max(\boldsymbol{\sigma}_p,\varepsilon)\right)\in\mathbb R^d,
\label{eq:final-stop-embedding}
\end{equation}

where $\boldsymbol{\mu}_p$ and $\boldsymbol{\sigma}_p$ are the featurewise mean and standard deviation of the PCA coordinates. The proposed controller uses $d=8$ and $c=3$.


The semantic representations are constructed in a shared feature space so that stops from different routes can be compared using the same coordinate system. This design allows a policy trained on one route to receive semantically consistent stop descriptors when applied to another route. Once constructed, the stop vectors are fixed throughout training and evaluation and are used as part of the local observation at each holding decision.

\subsection{Semantic-Enhanced RL Holding Policy}
\label{subsec:dqn-policy}

The semantic-enhanced observation defined in Equation~\eqref{eq:semantic-observation} is mapped to holding actions using a deep Q-learning policy shared by all buses. Parameter sharing allows decision experiences collected from different buses and stops to update a common action-value function $Q_\theta(\mathbf{o},a)$.
The Q-network $F_\theta$ maps an observation to the action values associated with the $A$ candidate holding durations:

\begin{equation}
\mathbf q_\theta(\mathbf{o})=F_\theta(\mathbf{o})\in\mathbb R^A,\qquad Q_\theta(\mathbf{o},a)=\left[\mathbf q_\theta(\mathbf{o})\right]_a.
\label{eq:q-network}
\end{equation}

For the proposed policy, the semantic stop vector $\mathbf z_i$ is concatenated directly with the dynamic spacing variables before being passed to the network. For the stop-ID benchmark, the categorical stop index is first mapped to a learned embedding. All observation designs otherwise use the same policy-learning procedure.
Each transition connects two consecutive decision epochs of the same bus and is stored in the replay buffer as

\begin{equation}
e_m=\left(\mathbf{o}_m,a_m,r_m,\mathbf{o}'_m\right).
\label{eq:dqn-transition}
\end{equation}

For transition $m$, the temporal-difference target is computed as

\begin{equation}
y_m=r_m+\gamma\,\operatorname{sg}\!\left[\max_{a'\in\mathcal A}Q_\theta(\mathbf{o}'_m,a')\right],
\label{eq:dqn-target}
\end{equation}

where $\operatorname{sg}[\cdot]$ denotes the stop-gradient operation. The same online Q-network is used to evaluate both the current and next observations, while the bootstrap target is treated as constant during backpropagation.

The network parameters are updated by minimizing the mean squared temporal-difference error over a minibatch $\mathcal B_m$:

\begin{equation}
\mathcal L(\theta)=\frac{1}{|\mathcal B_m|}\sum_{m\in\mathcal B_m}\left(Q_\theta(\mathbf{o}_m,a_m)-y_m\right)^2.
\label{eq:dqn-loss}
\end{equation}

During training, actions are selected using an $\epsilon$-greedy policy. During evaluation, the controller selects the action with the highest estimated value:

\begin{equation}
a_n^b=\arg\max_{a\in\mathcal A}Q_\theta(\mathbf{o}_n^b,a).
\label{eq:greedy-action}
\end{equation}

\section{Experimental Design}
\label{sec:experiment-setup}
\begin{figure}[htbp]
  \centering
  \includegraphics[width=\linewidth]{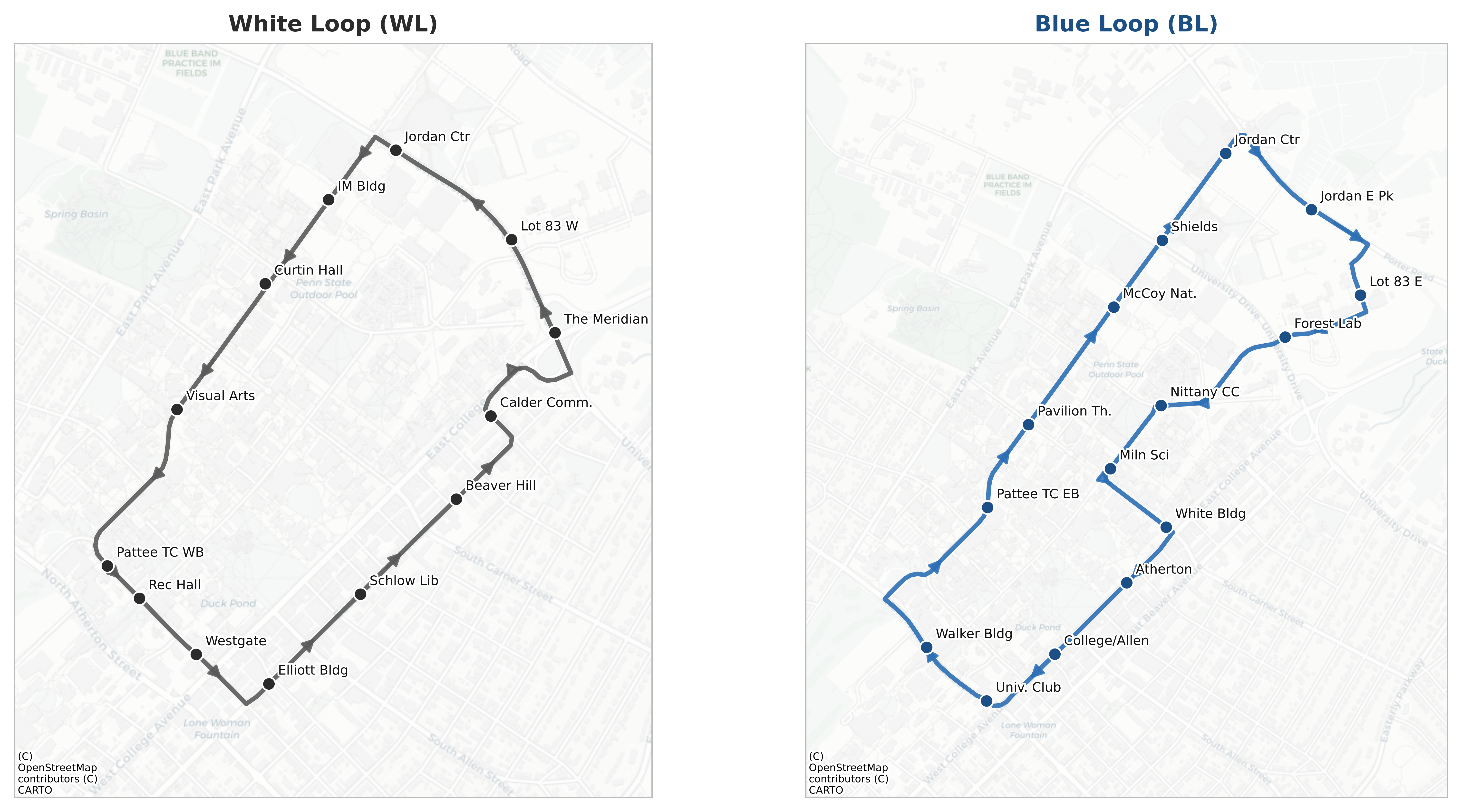}
  \caption{Route maps of the White Loop (WL) and Blue Loop (BL). Stop labels indicate stop names, and arrows indicate the direction of travel.}
  \label{fig:route-maps}
\end{figure}

This section describes the simulation environment, data preparation, control models, and evaluation settings used to investigate two research questions: (1) whether incorporating semantic stop information improves the same-route performance of an RL-based holding policy relative to benchmark control models, and (2) whether a policy learned on a source route can be effectively transferred to a target route. And it is examined through a cross-route experiment involving three policy training and deployment settings: zero-shot transfer without target-route training, warm-start fine-tuning initialized from the source-route checkpoint, and cold-start training from random initialization. 
The experiments use two Pennsylvania State University campus shuttle routes operating in State College, Pennsylvania, USA: the White Loop (WL) and Blue Loop (BL). In the cross-route experiment, WL serves as the source route and BL as the target route. Figure~\ref{fig:route-maps} shows the stop locations and directions of travel for both routes.

\subsection{Simulation Environment}\label{sec:simulation-environment}
Each simulation episode represents two hours of bus operation at a one-second time resolution, with buses initialized evenly along the route at the start of each episode. Passenger boarding and alighting occur at constant per-second service rates, and each stop enforces a fixed number of loading berths under a first-in-first-out, no-overtaking queue discipline. Stochastic passenger demand and link travel times are generated from route-specific parameters estimated from the historical data described in Section~\ref{sec:data-preparation}. When holding control is enabled, the controller is activated once a bus completes passenger service at a stop and selects a holding duration from a discrete action set before departing for the downstream link. Table~\ref{tab:simulator-settings} lists the exact simulation and control parameter values used throughout the experiments.

\begin{table}[!ht]
  \centering
  \caption{Simulator and control settings.}
  \label{tab:simulator-settings}
  \small
  \begin{tabular}{ll}
    \hline
    Setting & Value \\
    \hline
    Simulation duration & 7200s per episode \\
    Simulation step & 1 s \\
    Fleet size & 5 buses \\
    Vehicle capacity & 80 passengers \\
    Boarding / alighting rates & 0.8 / 1.0 passengers per second \\
    Stop berths & 6 berths per stop \\
    Stop queue discipline & FIFO, no overtaking \\
    Passenger arrivals & Poisson process (Sec.~\ref{sec:data-preparation}) \\
    Link travel times & Lognormal (Sec.~\ref{sec:data-preparation}) \\
    Holding action set & 0, 20, 40, and 60 s \\
    Evaluation seeds & 1--10 for the reported comparisons \\
    Validation seeds & 1001--1010 during model selection \\
    \hline
  \end{tabular}
\end{table}

\subsection{Data}
\label{sec:data-preparation}
Two categories of data are used in the experiment: (1) transit operational data used to construct and calibrate the simulation environment, and (2) stop-level contextual data used to construct the semantic representations introduced in Section~\ref{subsec:semantic-stop-info}.

\textbf{Transit operational data.} The simulator inputs are derived from historical CATA automatic vehicle location (AVL) records collected from August~15 to December~15, 2017, retaining daytime observations (8:00--20:00) after removing records with missing or invalid operational information. These records are used to estimate stop-specific Poisson passenger-arrival rates, gravity-model alighting probabilities incorporating walking-distance impedance, and link-specific lognormal travel-time distributions fitted from consecutive vehicle observations. Table~\ref{tab:route-input-summary} summarizes the resulting route-level inputs: compared with WL, BL has more stops, a longer route, and greater link travel-time variability, while WL has a slightly higher total arrival rate — differences that provide distinct operating conditions for same-route control and cross-route transfer evaluation.

\begin{table}[!ht]
  \centering
  \caption{Route-specific simulator inputs estimated from the AVL records.}
  \label{tab:route-input-summary}
  \small
  \begin{tabular}{lrrrrrrr}
    \hline
    Route & Stops & Length & Sum of mean & Mean link & Total arrival & Stop arrival & Demand \\
          &       & (km)   & link tt (min) & CV & rate (pax/min) & rate range (pax/min) & obs. \\
    \hline
    WL & 13 & 5.08 & 14.3 & 0.832 & 5.963 & 0.088--0.875 & 71,071 \\
    BL & 15 & 6.65 & 17.2 & 0.913 & 5.286 & 0.077--0.771 & 70,981 \\
    \hline
  \end{tabular}
\end{table}



\textbf{Semantic stop data.} The semantic stop data characterize the physical, functional, and operational context of each stop on the WL (13 stops) and BL (15 stops) routes: stop name, curbside type, route-context attributes, distances to adjacent stops, nearby points of interest (POIs), and AVL-derived demand descriptors (mean arrival rate, boarding variability, mean headway, and sample size). Nearby POIs are retrieved from OpenStreetMap via the Overpass API, mapped to a predefined campus land-use vocabulary (academic buildings, residence halls, libraries, student centers, athletic facilities, event venues, administrative facilities, parking areas, landmarks), with walking times estimated at a pedestrian speed of 1.35~m/s.
For instance, the WL stop \emph{Pattee TC WB} is a transit-center stop serving three routes (WL, WE, RL) at sequence position 6 of 13. It is located approximately one minute on foot from Pattee Library (library), three minutes from the West Halls residence complex (residence hall), and six to eight minutes from Old Main (administrative building), the Willard Building (academic building), and the HUB-Robeson Center (student center). The stop also exhibits relatively high and variable passenger demand, with a mean arrival rate of 0.69~pax/min, a boarding coefficient of variation of 0.93, and 6{,}871 valid observations. These collected facts serve as inputs to the semantic representation procedure described in Section~\ref{subsec:semantic-stop-info}.

\subsection{Benchmark Control Models}
\label{sec:control-methods}
The proposed \emph{Spacing + semantic} RL policy combines normalized spacing features with the semantic stop representation described in Section~\ref{subsec:semantic-stop-info}. It is compared with four benchmark control models. The first is the \emph{no-control} baseline, which applies no holding intervention and allows buses to depart immediately after completing passenger service. The second is a calibrated rule-based implementation of the headway-based holding model proposed by Daganzo \cite{Daganzo2009Headway}, with the control gain $\alpha$ and slack fraction $f$ selected through grid search and the maximum holding time fixed at 60~s. The other two benchmarks are RL-based models that share the same network architecture, action space, and training procedure as the proposed policy but use different state representations. The \emph{Spacing} model uses only normalized forward and backward spatial spacings, whereas the \emph{Spacing + stop ID} model augments these spacing features with a learned categorical stop embedding.

\subsection{Model Training and Evaluation}
\label{sec:evaluation-protocol}
The no-control and rule-based benchmarks are evaluated directly, whereas each RL controller is trained for 200 episodes, validated every 10 episodes, and selected by the checkpoint with the lowest mean validation headway variation (ties broken by total bunching count). Final evaluation uses independent seeds and deterministic greedy actions. All methods are compared on four metrics: headway variation (service regularity), total bunching count, average holding time (control intervention), and average passenger waiting time. Lower values are preferred throughout, though holding time is treated as an operational cost to be weighed against gains in the other three metrics. Space-time diagrams are used only for visual and diagnostic analysis.

\textbf{Performance metrics.} Four performance metrics are reported for each evaluation episode. \textbf{Bunching count} is defined as the total number of stop-level bunching detections over the episode. A bunching detection occurs when at least two buses are simultaneously present at the same stop. \textbf{Headway variation} is the standard deviation of the actual departure headways observed across bus departure events. \textbf{Average waiting time} is the cumulative passenger waiting time in all stop queues divided by the total number of passenger arrivals. \textbf{Average holding time} is the mean selected holding duration across all holding decisions, including zero-second holding actions.


\section{Results}
\label{sec:results}
This section evaluates within-route control performance, cross-route policy reuse, and the behavioral patterns induced by the learned policies.

\subsection{Within-Route Performance}
Table~\ref{tab:wl-results} compares the control methods using the four previously introduced performance metrics. First, all three RL controllers achieve lower mean headway variation, fewer bunching events, and shorter passenger waiting times than both benchmark methods—the Daganzo rule and the no-control baseline. These findings suggest that the learned policies adapt more effectively to the stochastic and state-dependent operating conditions represented in the simulation.
Second, the comparison among the RL state representations highlights the value of semantic stop information. Adding a categorical stop identifier does not improve the spacing-only policy in terms of headway variation, bunching count, or passenger waiting time. This finding is consistent with prior evidence that arbitrary stop identifiers provide limited additional information beyond compact spacing-based states \cite{XuEtAl2025Systematic}. In contrast, augmenting the state with semantic stop information improves performance across multiple dimensions.
Third, the results reflect the multi-objective nature of the control problem, which requires balancing service regularity, passenger delay, and control effort. Among the RL controllers, our semantic policy achieves the lowest mean headway variation, passenger waiting time, and holding time, whereas the spacing-only policy produces marginally fewer bunching events at the cost of substantially longer holding. Overall, the semantic policy provides the most favorable balance across the evaluated metrics.

\begin{table}[!ht]
  \caption{Within-Route Performance}\label{tab:wl-results}
  \begin{center}
  \small
  \begin{tabular}{lrrrr}
    \hline
    Method & Headway var. (s) $\downarrow$ & Bunching $\downarrow$ & Hold (s) $\downarrow$ & Wait (min) $\downarrow$ \\
    \hline
    Do-nothing & $201.5\pm39.2$ & $66.0\pm16.3$ & 0.0 & $3.85\pm1.1$ \\
    Daganzo rule ($\alpha=0.1$, $f=0.1$) & $189.2 \pm 33.3$ & $31.5 \pm 9.4$ & $\mathbf{12.2 \pm 1.1}$ & $3.38 \pm 0.8$ \\
    RL: spacing & $140.2\pm32.3$ & $\mathbf{8.0\pm5.3}$ & $32.4\pm1.4$ & $2.92\pm0.5$ \\
    RL: spacing + stop id & $157.0\pm53.8$ & $15.1\pm10.0$ & $27.0\pm2.1$ & $2.99\pm0.7$ \\
    RL: spacing + semantic & $\mathbf{128.7\pm21.4}$ & $9.7\pm3.8$ & $25.1\pm1.0$ & $\mathbf{2.57\pm0.3}$ \\
    \hline
  \end{tabular}
  \end{center}
\end{table}

\subsection{Cross-Route Performance}
\begin{figure}[htbp]
  \centering
  \includegraphics[width=\linewidth]{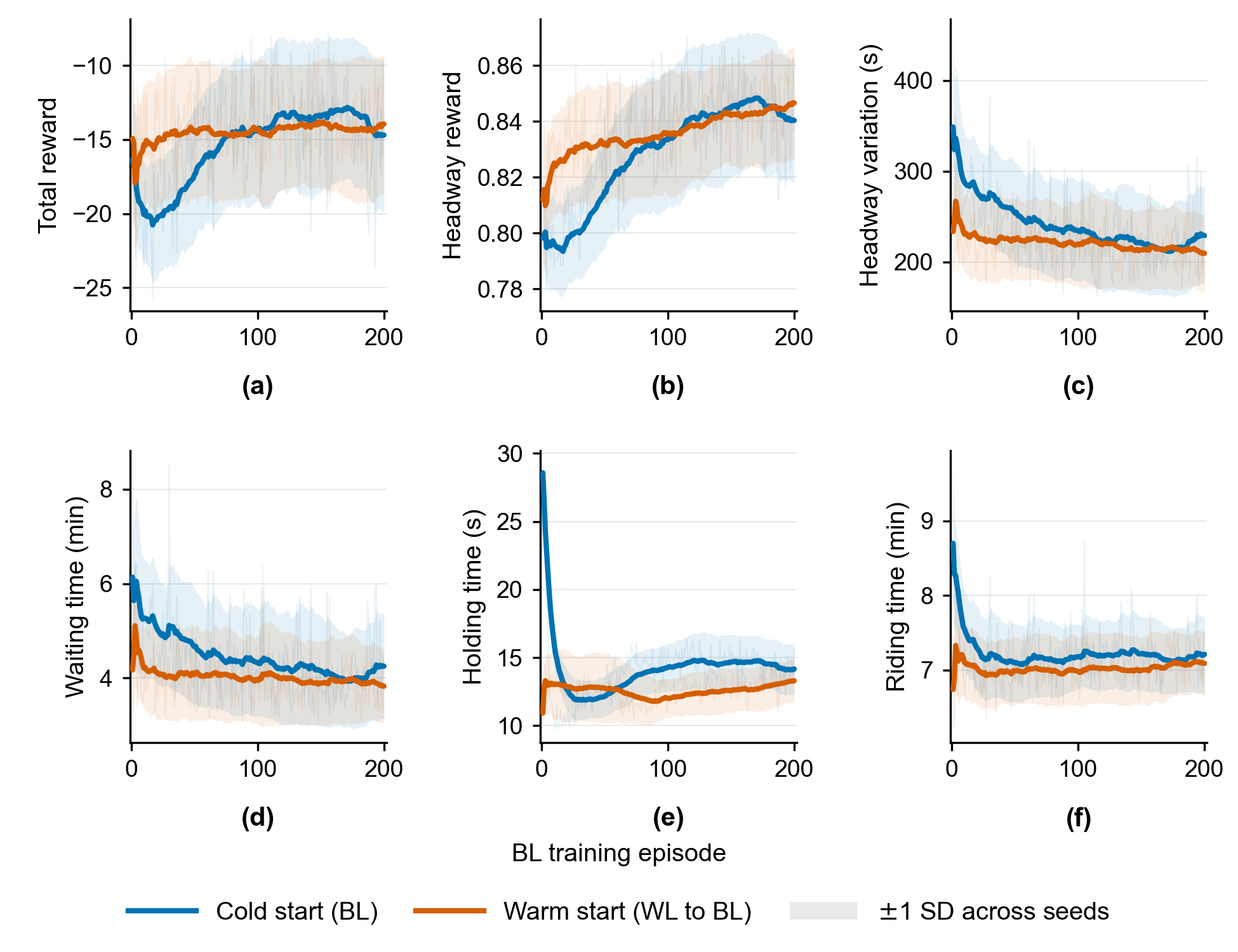}
  \caption{Cold-start and warm-start training trajectories on the target route. Panels show total reward, headway reward, headway variation, passenger waiting time, holding time, and passenger riding time. Thin lines denote seed-level episode means, and shaded bands indicate $\pm 1$ standard deviation across five seeds.}\label{fig:transfer-reward}
\end{figure}

Overall, these results support the feasibility of semantic-enhanced RL policy transfer across routes. Figure~\ref{fig:transfer-reward} first examines the training behavior of the semantic-enhanced RL policy by comparing cold-start training with warm-start fine-tuning. The training curves show that the warm-start policy begins from a better initial performance level, with higher total reward and headway reward, as well as lower headway variation and passenger waiting time during the early training episodes. It also reaches a stable performance range more quickly. These trajectories suggest that the policy learned on the source route contains transferable information that can be reused on the target route, thereby reducing the amount of target-route training needed to obtain a reasonable control policy. Table~\ref{tab:transfer-results} reports the final target-route control performance evaluated over several random seeds. Cold-start training achieves the lowest headway variation, bunching count, and passenger waiting time among the RL variants. This result is expected because the policy is trained entirely on the target route and can therefore specialize to route-specific dynamics. However, this performance is achieved with greater holding effort than the transferred policies. In contrast, zero-shot transfer improves bunching count and passenger waiting time relative to the no-control baseline, but provides little improvement in headway variation, indicating that direct policy reuse is feasible but insufficient without target-route adaptation. After target-route adaptation (warm-start with fine-tuning), the transferred policy further improves over zero-shot transfer and reaches a more balanced performance level. Notably, it requires the lowest holding time among the active controllers, suggesting that warm-start fine-tuning can adapt the source-route policy to the target route while maintaining a lower control effort. Together with the faster early-stage learning shown in Figure~\ref{fig:transfer-reward}, these results indicate that source-route knowledge can reduce target-route training effort and support effective cross-route policy adaptation.

\begin{table}[htbp]
  \caption{Transfer Performance for the Spacing+Semantic RL Policy}\label{tab:transfer-results}
  \begin{center}
  \small
  \begin{tabular}{lrrrr}
    \hline
    Method & Headway var. (s) $\downarrow$ & Bunching $\downarrow$ & Hold (s) & Wait (min) $\downarrow$ \\
    \hline
    Do-nothing & $255.4\pm36.5$ & $61.7\pm8.7$ & 0.0 & $4.91\pm0.9$ \\
    Daganzo rule ($\alpha=0.1$, $f=0.1$) & $262.1 \pm 78.2$ & $28.2$ & $16.4$ & $4.67$ \\
    Zero-shot (from source-route) & $242.7 \pm 35.4$ & $33.4 \pm 8.4$ & $17.1$ & $4.35 \pm 0.84$ \\
    Warm-start (fine-tuned) & $223.7\pm75.4$ & $26.0\pm11.8$ & $\mathbf{16.0\pm1.2}$ & $4.21\pm1.6$ \\
    Cold-start (from scratch) & $\mathbf{201.2\pm68.0}$ & $\mathbf{18.2\pm12.3}$ & $22.4\pm3.3$ & $\mathbf{3.79\pm1.0}$ \\
    \hline
  \end{tabular}
  \end{center}
\end{table}

\subsection{Bunching Patterns and Holding Actions}
\begin{figure}[htbp]
  \centering
  \includegraphics[width=0.99\linewidth]{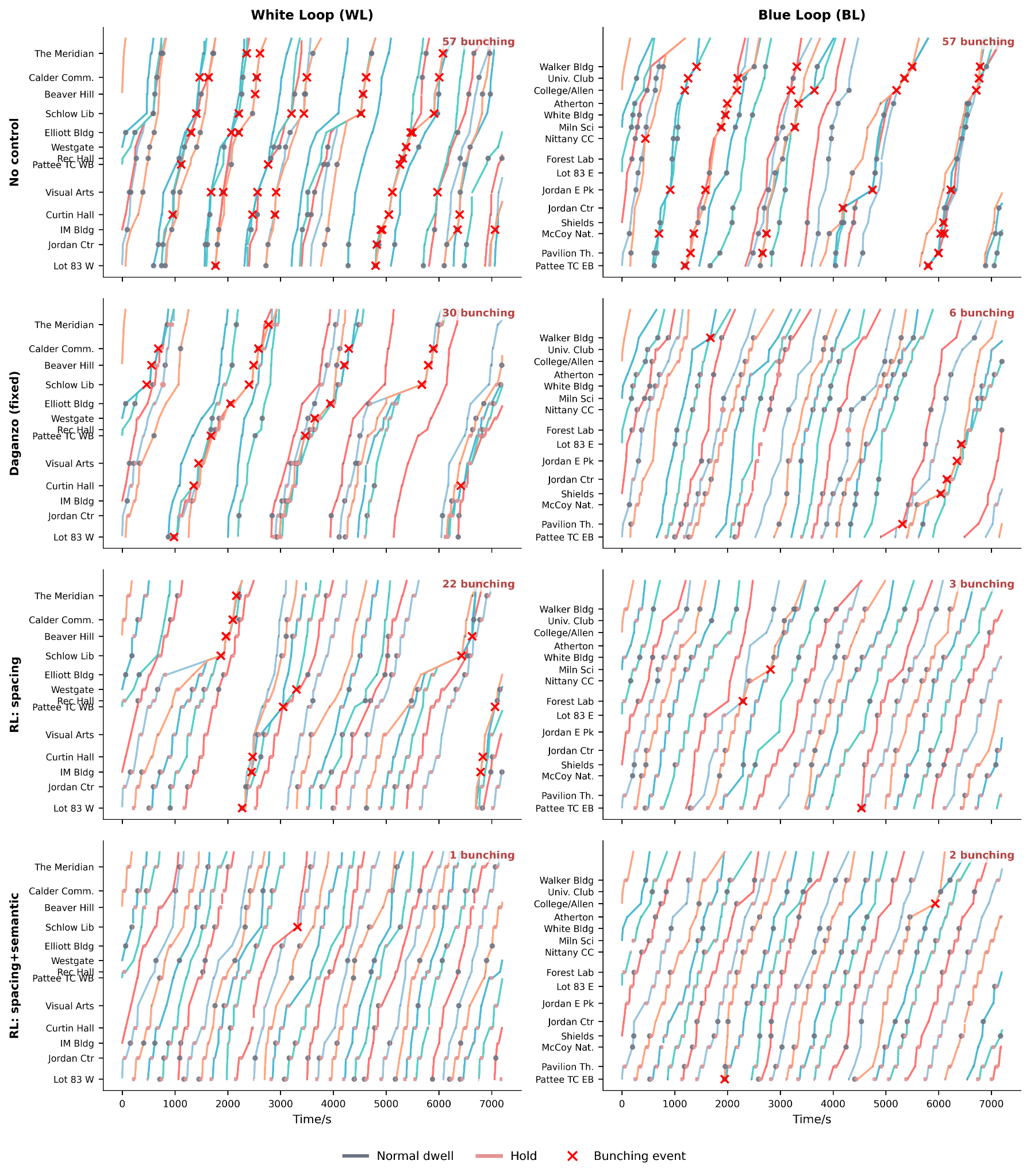}
  \caption{Representative simulated space-time trajectories for WL and BL under no control, best-tuned Daganzo rule, spacing-only RL, and semantic-enhanced RL. Colored lines denote vehicle trajectories, gray and pink horizontal segments denote normal dwell and holding, respectively, and red crosses denote detected bunching events. All panels use the same evaluation seed for illustration.}\label{fig:wl-spacetime-results}
\end{figure}

Figure~\ref{fig:wl-spacetime-results} complements the aggregate metrics by showing representative simulated space--time diagrams for both routes. For each route, all control methods are evaluated under the same random seed and route configuration; therefore, the diagrams are intended to illustrate temporal control behavior rather than provide a statistical ranking. Red crosses indicate the stop--time locations where bunching events are detected, showing not only how many bunching events occur but also where they emerge and propagate along the route. In the no-control cases, bunching events are widely distributed across both routes and often appear repeatedly over consecutive stops, suggesting that once headway disturbances form, they can persist and propagate downstream without intervention. These locations can be interpreted as less resilient portions of the route in this realization, where the service has limited ability to recover naturally from headway deviations. With active control, the number of bunching events decreases and the remaining events become more localized, indicating that holding actions can interrupt the propagation of bunching and improve route-level resilience. In the WL realization, the bunching count decreases from 57 under no control to 30 under the Daganzo rule, 22 under spacing-only RL, and 1 under spacing+semantic RL. The corresponding counts on BL are 57, 6, 3, and 2. The increasingly clear separation between vehicle trajectories further illustrates how active holding can maintain more regular service over time.

\begin{figure}[htbp]
  \centering
  \includegraphics[width=\linewidth]{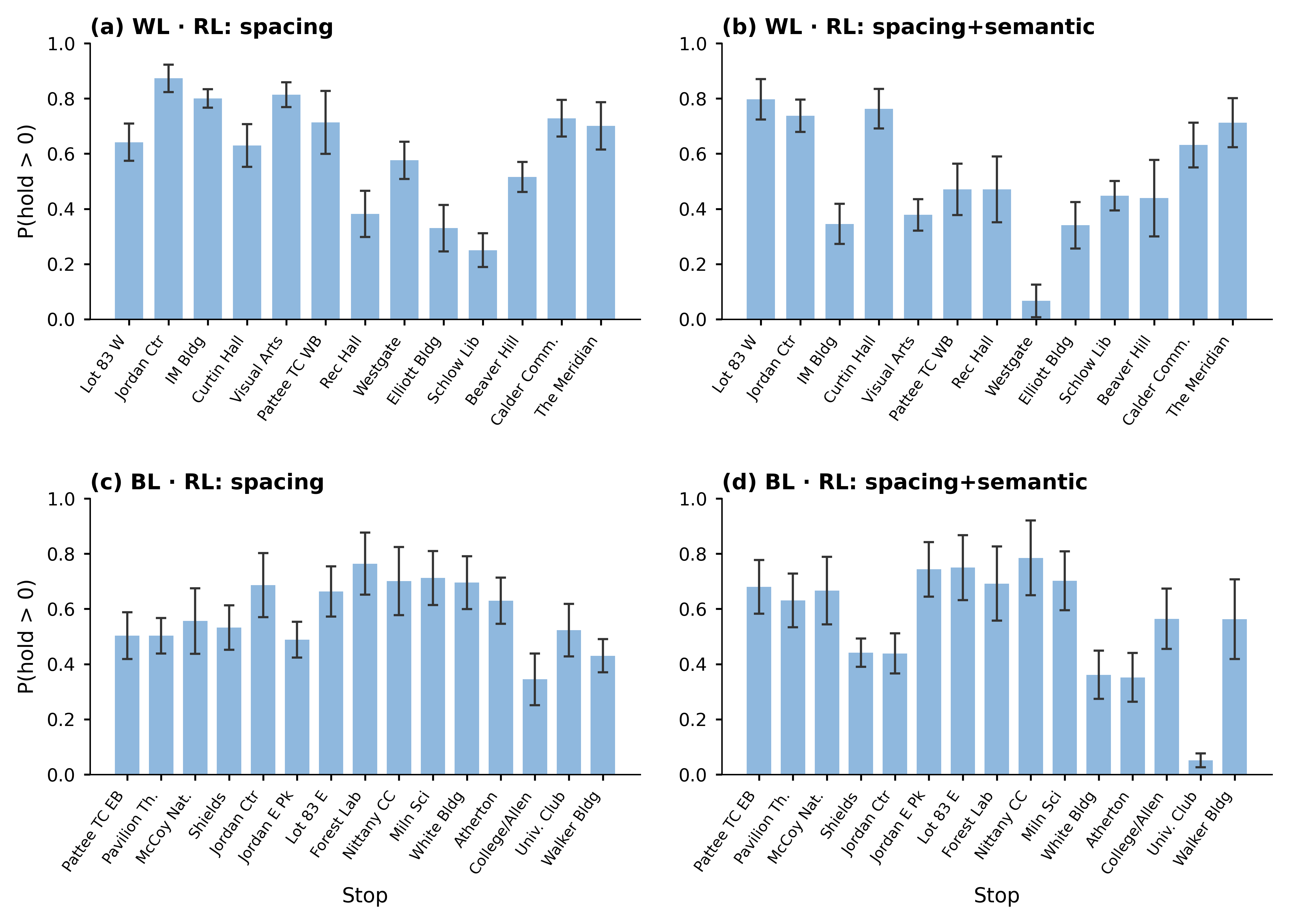}
  \caption{Stop-level probability of selecting a positive holding action for (a) WL spacing-only RL, (b) WL spacing+semantic RL, (c) BL spacing-only RL, and (d) BL spacing+semantic RL. Bars show the mean of the per-seed positive-hold probability across ten evaluation seeds, and error bars indicate $\pm 1$ standard deviation. A positive holding action has a selected holding time greater than zero.}\label{fig:hold-probability-by-stop}
\end{figure}

Figure~\ref{fig:hold-probability-by-stop} compares the stop-level hold probabilities of the spacing-only RL policy and the semantic-enhanced RL policy on the WL and BL routes. The nonuniform profiles show that both policies learn stop-specific control patterns rather than applying holding uniformly along the route. Adding semantic stop information further changes where holding is applied: the unweighted stop-level average hold probability decreases from 0.612 to 0.508 on WL and from 0.582 to 0.561 on BL, while the locations with increased or reduced holding also shift. For example, the semantic-enhanced RL policy substantially reduces holding at Westgate on WL and University Club on BL, while retaining or increasing holding at Lot~83~W on WL and Lot~83~E on BL. These changes suggest that semantic stop information affects the spatial allocation of control effort and may help the policy identify more suitable locations for regulating downstream service. Several of these stops correspond to recognizable campus activity or access points, making their role as potential control locations plausible. 

\section{Conclusion}\label{sec:conclusion}

This paper develops and evaluates a language-model-assisted semantic stop representation for reinforcement-learning-based bus holding control. The representation complements dynamic operational states by capturing the functional context and recurring operational characteristics that differentiate individual stops. It is constructed from stop attributes, surrounding activity context, historical demand and service summaries, and schema-constrained LLM outputs. The resulting representation is integrated into an event-driven RL controller. The simulation results show that the semantic-enhanced controller provides the most favorable overall balance among the evaluated RL state designs in the same-route experiment. It achieves the lowest headway variation, passenger waiting time, and holding time among the RL controllers, while attaining the second-lowest bunching count. Stop-level action patterns further suggest that semantic information changes the spatial allocation of holding effort rather than simply increasing holding intensity. In the cross-route experiment, zero-shot deployment of the source-route policy provides partial benefits but does not fully adapt to the target route. Fine-tuning improves the transferred policy and provides better initialization and early-stage learning, whereas cold-start training on the target route achieves the strongest final regularity and passenger-delay performance after convergence.

Overall, the findings indicate that semantic stop representations can complement conventional bus holding states and support source-to-target policy reuse and adaptation in the studied setting. From an implementation perspective, the representations are constructed and cached offline, so language-model inference, text embedding, and representation alignment do not add computation or latency to the real-time control loop. Representing stops in a shared feature space also provides a practical basis for reusing learned policies across related routes. Such reuse may reduce target-route training and retraining needs. It may also enable agencies to test, update, and deploy control strategies more rapidly than developing a separate policy from scratch for each route. However, the study is limited to simulation-based experiments on two university shuttle routes and one primary transfer direction. Future work should evaluate the framework on larger and more heterogeneous transit networks, quantify the data and computational savings associated with policy reuse, examine additional transfer directions, test robustness under demand shifts and disruptions, and further isolate the effects of semantic labels from those of the underlying operational information.


\section*{GENERATIVE AI DISCLOSURE}
\begin{sloppypar}
ChatGPT (OpenAI) was used to assist with language editing and
rephrasing portions of the manuscript. The authors reviewed and
verified all generated text and take full responsibility for the
content of the manuscript.
\end{sloppypar}

\section*{AUTHOR CONTRIBUTIONS} 
All authors contributed to the study conception and design, data collection, methodology development, analysis and interpretation of the results, and manuscript preparation and revision. All authors reviewed the results and approved the final version of the manuscript.

\section*{DECLARATION OF CONFLICTING INTERESTS}


The authors declared no potential conflicts of interest with respect to the research, authorship, and/or publication of this article.

\section*{FUNDING}

The authors disclosed no financial support for the research, authorship, and/or publication of this article.

\newpage
\bibliographystyle{trb}
\bibliography{trb_template}
\end{document}